\documentclass[letterpaper]{article} 
\usepackage{aaai25}  
\usepackage{times}  
\usepackage{helvet}  
\usepackage{courier}  
\usepackage[hyphens]{url}  
\usepackage{graphicx} 
\usepackage{natbib}  
\usepackage{caption} 
\usepackage{amsmath}
\usepackage{multirow}
\usepackage{algorithm}
\usepackage{algorithmic}

\usepackage{newfloat}
\usepackage{listings}
\DeclareCaptionStyle{ruled}{labelfont=normalfont,labelsep=colon,strut=off} 
\floatstyle{ruled}
\newfloat{listing}{tb}{lst}{}
\floatname{listing}{Listing}
\title{Hierarchical Document Parsing via Large Margin Feature Matching and Heuristics}
\author {
    Duong Anh Kiet
}
\affiliations{
    L3i Laboratory, La Rochelle University\\
    17042 La Rochelle Cedex 1 - France\\
    anh.duong@univ-lr.fr
}

\begin{document}

\maketitle

\begin{abstract}
We present our solution to the AAAI-25 VRD-IU challenge, achieving first place in the competition. Our approach integrates large margin loss for improved feature discrimination and employs heuristic rules to refine hierarchical relationships. By combining a deep learning-based matching strategy with greedy algorithms, we achieve a significant boost in accuracy while maintaining computational efficiency. Our method attains an accuracy of 0.98904 on the private leaderboard, demonstrating its effectiveness in document structure parsing.
\end{abstract}

\begin{links}
    \link{Code}{https://github.com/ffyyytt/VRUID-AAAI-DAKiet}
    \link{Models}{https://www.kaggle.com/models/fdfyaytkt/vruid-aaai-dakiet/PyTorch/figure}
\end{links}

\section{Introduction}
Document structure parsing is essential for visually rich document understanding, enabling applications such as information retrieval and document summarization~\cite{xu2020layoutlm}. Unlike traditional text-based document processing, mining reports and other complex documents contain heterogeneous semantic entities, requiring hierarchical parsing to infer parent-child relationships~\cite{li2020docbank}.

The \textit{AAAI-25 Visually-Rich Document (VRD-IU) Leaderboard} task involves predicting parent-child relationships among detected entities in mining reports. This problem is challenging due to scanned and inconsistently structured documents, necessitating inference from spatial and contextual cues. Performance is evaluated using accuracy.

In this paper, we present our solution, which ranked first in the competition. We combines large margin loss for improved feature discrimination with heuristic greedy algorithms for efficient hierarchical assignment. Method~\ref{sec:method} details our approach, Experiments~\ref{sec:experiments} presents experimental results, and Conclusion section~\ref{sec:conclusion} concludes with insights and future directions.  
\section{Related Work}\label{sec:related}
Documents inherently exhibit a hierarchical structure, where parent-child relationships are fundamental across various formats, such as figure-caption and section-paragraph associations in reports \cite{ding2024deep}. Accurately modeling these relationships is crucial for document structure parsing. Prior works such as DocStruct \cite{wang2020docstruct}, SERA \cite{zhang2021entity}, and KVPFormer \cite{hu2023question} have significantly contributed to this task by leveraging deep learning models for entity linking. However, these methods rely on conventional distance metrics, which may struggle to differentiate entities with highly similar feature representations \cite{duong2025scalable}.

Recent advancements in matching take inspiration from CLIP \cite{radford2021learning}, instead of cross-modal learning between text and images, document structure parsing deals with entities of the same modality. However, we also need feature representations for child $fc$ and for parent $fp$. In terms of feature separation, large margin loss functions such as LMCot~\cite{duong2022large}, ArcFace~\cite{deng2019arcface}, and NormFace~\cite{wang2017normface} improve feature extraction by enforcing larger inter-class separability while maintaining intra-class compactness.

Beyond deep learning-based approaches, heuristic rules and greedy algorithms have proven effective in competitive settings\cite{duong2025addressing}. In the next section, we detail our integration of large margin loss for matching and the heuristic rules applied in the competition.  
\section{Method}
In this section we introduce our loss function with improvements from large margin loss. And then the heuristic rules we observed to increase the performance of the model.
\label{sec:method}
\subsection{Margin loss for matching}
Starting with the loss function in CLIP \cite{radford2021learning}:
\begin{equation}
\label{eq:clip}
\small
    L_{\text{CLIP}}=-\frac{1}{N_c}\sum\limits_{i=1}^{N}{\log }\frac{ e^{ fc_{i} \times  fp_{y_i}^{T} }}{\sum\limits_{j=1}^{N_p}{{e}^{fc_{i} \times fp_{j}^{T} }}},
\end{equation}
where \( N_c \) and \( N_p \) denote the number of child entities requiring a parent and the number of potential parent entities, respectively. The feature representations of child and parent entities are denoted as \( fc_i \) and \( fp_i \), where \( i \) is the index of the entity. The term \( fp_{l_i} \) corresponds to the feature of the actual parent of the \( i \)-th child entity, where \( y_i \) indicates its parent’s index. In other words, the numerator captures the similarity between a child and its correct parent, while the denominator sums over all potential parent entities.

Then we can reformulate the equation \ref{eq:clip} as:
\begin{equation}
\label{eq:clip2}
\small
    L_{\text{CLIP}}=-\frac{1}{N_c}\sum\limits_{i=1}^{N}{\log }\frac{ e^{ 
 s\left\|fc_{i}\right\| \left\|fp_{y_i}^{T}\right\|  \cos\left( \theta_{i,y_i} \right) }}{\sum\limits_{j=1}^{N_p}{{e}^{ s\left\|fc_{i}\right\|\left\|fp_{j}^{T}\right\| \cos\left( \theta_{i,j} \right) }}},
\end{equation}
where $\theta_{i,j}$ is the angle between the child feature vector $fc_{i}$ and the parent feature vector $fp_{j}$. A scaling factor \( s \) is then introduced to restore the magnitude after normalization.

Following NormFace \cite{wang2017normface}, we normalize the feature vectors so that \( \left\|fc_{i}\right\|=\left\|fp_{j}\right\|=1 \) for all \( i, j \). Building upon ArcFace \cite{deng2019arcface}, we further introduce a margin parameter \( m \), which strengthens the relationship between parent-child pairs while increasing the separation between unrelated entities. By this way, the loss function is rewritten as:

\begin{equation}
\label{eq:loss}
\small
    L_{\text{matching}}=-\frac{1}{N_c}\sum\limits_{i=1}^{N}{\log }\frac{ e^{ s \times \cos\left( \theta_{i,y_i} + m \right) }}{e^{ s \times \cos\left( \theta_{i,y_i} + m \right) } + \sum\limits_{j=1, j\neq i}^{N_p}{{e}^{ s \times \cos\left( \theta_{i,j}\right) }}},
\end{equation}

Thus, we effectively incorporate advancements in large margin cosine loss into feature matching learning. With this loss function, the feature extraction process is optimized such that, even with the margin parameter \( m \), features of related parent-child entities remain closer in the learned space compared to unrelated ones. This encourages the model to form more discriminative representations, enhancing the accuracy of hierarchical relationship inference.

\subsection{Greedy algorithms}
In addition to the loss function introduced earlier, we identified several structural patterns that improve both the accuracy and computational efficiency of our approach. These heuristics allow for a more efficient assignment of parent-child relationships, reducing the need for exhaustive pairwise comparisons.
First, entities belonging to specific categories never have a parent, including:  \texttt{abstract}, \texttt{appendix\_list}, \texttt{cross}, \texttt{figure}, \texttt{form\_title}, \texttt{list\_of\_figures}, \texttt{list\_of\_tables}, \texttt{other}, \texttt{references}, \texttt{report\_title}, \texttt{section}, \texttt{summary}, \texttt{table}, \texttt{table\_of\_contents}, and \texttt{title}.

Second, a strong hierarchical relationship exists among the categories:  
\texttt{section}, \texttt{subsection}, \texttt{subsubsection}, \texttt{subsubsubsection}, and \texttt{paragraph}. These entities follow a strict sequential structure, where each entity is assigned as a child to the nearest preceding entity in the order.

Third, specific entities exhibit predefined parent-child dependencies:  
\begin{itemize}
    \item \texttt{table\_caption} is always a child of \texttt{table}.  
    \item \texttt{figure\_caption} is always a child of \texttt{figure}.  
    \item \texttt{form} is always a child of \texttt{summary}, \texttt{abstract}, \texttt{section}, \texttt{subsection}, \texttt{subsubsection}, or \texttt{subsubsubsection}.  
    \item \texttt{list} is a child of \texttt{paragraph}, \texttt{section}, \texttt{subsection}, \texttt{subsubsection}, or \texttt{subsubsubsection}.  
    \item \texttt{form\_body} is a child of \texttt{form\_title}, \texttt{summary}, \texttt{abstract}, \texttt{section}, \texttt{subsection}, \texttt{subsubsection}, or \texttt{subsubsubsection}.  
\end{itemize}  
By enforcing these structured dependencies, we further improve the accuracy of entity relationship prediction while maintaining computational efficiency.
\section{Experiments}
In this section, we present the competition dataset, our implementation, and the archived results of the competition.
\label{sec:experiments}
\subsection{Dataset}
The dataset contains 571 training, 165 validation, and 81 test documents. Labels are provided only for the training set, while the validation and test sets are used for the public and private leaderboards. Each document contains multiple objects, each with a bounding box in the scanned document and optional text content. The task is to predict the parent object for each object, evaluated using accuracy.
\subsection{Implemtentation}
We use a pre-trained BART-large \cite{lewis2019bart} for text feature extraction and a pre-trained SwinV2-Tiny \cite{liu2022swin} for image features. The extracted features are concatenated with additional information such as page number and bounding box position. Two separate encoders, each with a 512-dimensional output, are used to extract features for parent and child entities.
\subsection{Results}
Due to time constraints in the competition, we were unable to conduct extensive experiments on alternative methods or additional datasets. Table~\ref{tab:result} presents the main results obtained on the competition dataset, evaluated on the validation set and test set.
\begin{table}[h]
\centering
\begin{tabular}{c|cc}
\multirow{2}{*}{Method}          & \multicolumn{2}{c}{Accuracy}                              \\ \cline{2-3} 
                                 & val                         & test                        \\ \hline
loss only                        & 0.79674                     & 0.85824                     \\
\multicolumn{1}{l|}{loss+greedy} & \multicolumn{1}{l}{0.97369} & \multicolumn{1}{l}{0.98904}
\end{tabular}
\caption{Performance comparison of our methods.}
\label{tab:result}
\end{table}

\section{Conclusion}\label{sec:conclusion}
In this work, we proposed a novel approach for document structure parsing, combining large margin loss with heuristic-based greedy algorithms. Our method enhances feature discrimination while efficiently capturing hierarchical relationships. The results demonstrate the effectiveness of integrating deep learning with rule-based refinements, achieving state-of-the-art performance in the AAAI-25 VRD-IU challenge. Future work includes exploring more diverse datasets and refining heuristic rules for better generalization.

\bibliography{references}

\end{document}